\documentclass{article}
\pdfoutput=1
\usepackage{arxiv}

\usepackage[utf8]{inputenc} 
\usepackage[T1]{fontenc}    
\usepackage{hyperref}       
\usepackage{url}            
\usepackage{booktabs}       
\usepackage{nicefrac}       
\usepackage{microtype}      
\usepackage{lipsum}		
\usepackage{graphicx}
\usepackage{algorithm}
\usepackage{algorithmic}
\usepackage{amsmath,amssymb,amsfonts}

\title{Neural Generators of Sparse Local Linear Models \\for Achieving both Accuracy and Interpretability}

\author{Yuya Yoshikawa \\
	STAIR Lab, Chiba Institute of Technology / RIKEN AIP\\
	\texttt{yoshikawa@stair.center} \\
	\And
	Tomoharu Iwata \\
	NTT Communication Science Laboratories\\
	\texttt{tomoharu.iwata.gy@hco.ntt.co.jp} \\
}

\newcommand{\argmax}{\mathop{\rm argmax}\limits}

\newcommand{\mat}[1]{\boldsymbol{#1}}
\newcommand{\bref}[1]{(\ref{#1})}
\newcommand{\set}[1]{\mathcal{#1}}

\renewcommand{\algorithmicrequire}{\textbf{Input:}}
\renewcommand{\algorithmicensure}{\textbf{Output:}}

\begin{document}
\maketitle

\begin{abstract}
For reliability, it is important that the predictions made by machine learning methods
are interpretable by human.
In general, deep neural networks (DNNs) can provide accurate predictions,
although it is difficult to interpret why such predictions are obtained by DNNs.
On the other hand, interpretation of linear models is easy,
although their predictive performance would be low since
real-world data is often intrinsically non-linear.
To combine both the benefits of the high predictive performance of DNNs and
high interpretability of linear models into a single model,
we propose neural generators of sparse local linear models (NGSLLs).
The sparse local linear models have high flexibility as they can approximate non-linear functions.
The NGSLL generates sparse linear weights for each sample using DNNs
that take original representations of each sample (e.g., word sequence) 
and their simplified representations (e.g., bag-of-words) as input.
By extracting features from the original representations,
the weights can contain rich information to achieve high predictive performance.
Additionally, the prediction is interpretable because it is obtained by the inner product between the simplified representations and the sparse weights, where only a small number of weights are selected by our gate module in the NGSLL. 
In experiments with real-world datasets, we demonstrate the effectiveness of the NGSLL
quantitatively and qualitatively by evaluating prediction performance
and visualizing generated weights on image and text classification tasks.
\end{abstract}


\section{Introduction}\label{sec:intro}
Thanks to the recent advances in machine learning research, the predictive performance by machine learning has continually improved, and the advances present results that machine learning has surpassed human's performance in some tasks in the field of computer vision and natural language processing~\cite{he2016,zhang2019}.
Therefore, machine learning has recently come into wide use in services and products.

Recently, in supervised learning, deep neural networks (DNNs) have been essential to achieving high predictive performance.
In general, the benefits of DNNs are: 1) they are able to automatically extract meaningful features from the original representation of inputs, for example, RGB values of pixels from images and sequences of words in documents; 2) they achieve high predictive performance using the extracted features; and 3) they connect the feature extraction and the prediction seamlessly and learn their parameters efficiently via backpropagation.
However, as DNNs generally have an extremely large number of parameters, and are made of multiple non-linear functions, it is difficult for humans to interpret how such predictions are performed by DNNs.

For reliability, it is important that the predictions made by machine learning methods are interpretable by humans, i.e., one can understand what features are important for the predictions and how the predictions are made.
Therefore, in recent years interpretable machine learning has been intensively studied in not only the machine learning community~\cite{wilson2017} but also in various other areas such as computer vision~\cite{escalante2018}, natural language processing~\cite{mathews2019}, as well as in materials and medical science~\cite{dimiduk2018,tjoa2019}.
The representative methods for interpretable machine learning are Local Interpretable Model-agnostic Explainations (LIME)~\cite{ribeiro2016} and SHapley Additive exPlanations (SHAP)~\cite{lundberg2017}.
However, as stated in the paper~\cite{ribeiro2018}, these methods have mainly two drawbacks: 1) pseudo-samples used for interpretable model learning might not be part of the original input distribution, and 2) the learning of the interpretable models is time-consuming because it is executed in the test phase.

Linear models with feature selection such as Lasso~\cite{tibshirani1996} are easy for humans to interpret their predictions, as each of their own weights indicates the importance of its corresponding feature.
However, in cases where data distribution is intrinsically non-linear, the predictive performance of the linear models would be low due to the low flexibility of their linear function.
Moreover, we sometimes need to transform the original representation
so that it is compatible with linear models.
For example, in text classification tasks,
text documents are originally represented by sequences of words with various length,
they are transformed to bag-of-words vectors with fixed length with vocabulary size.
This would remove important features included in the original representation.

\begin{figure*}[t]
\centering
\includegraphics[width=\textwidth,pagebox=artbox]{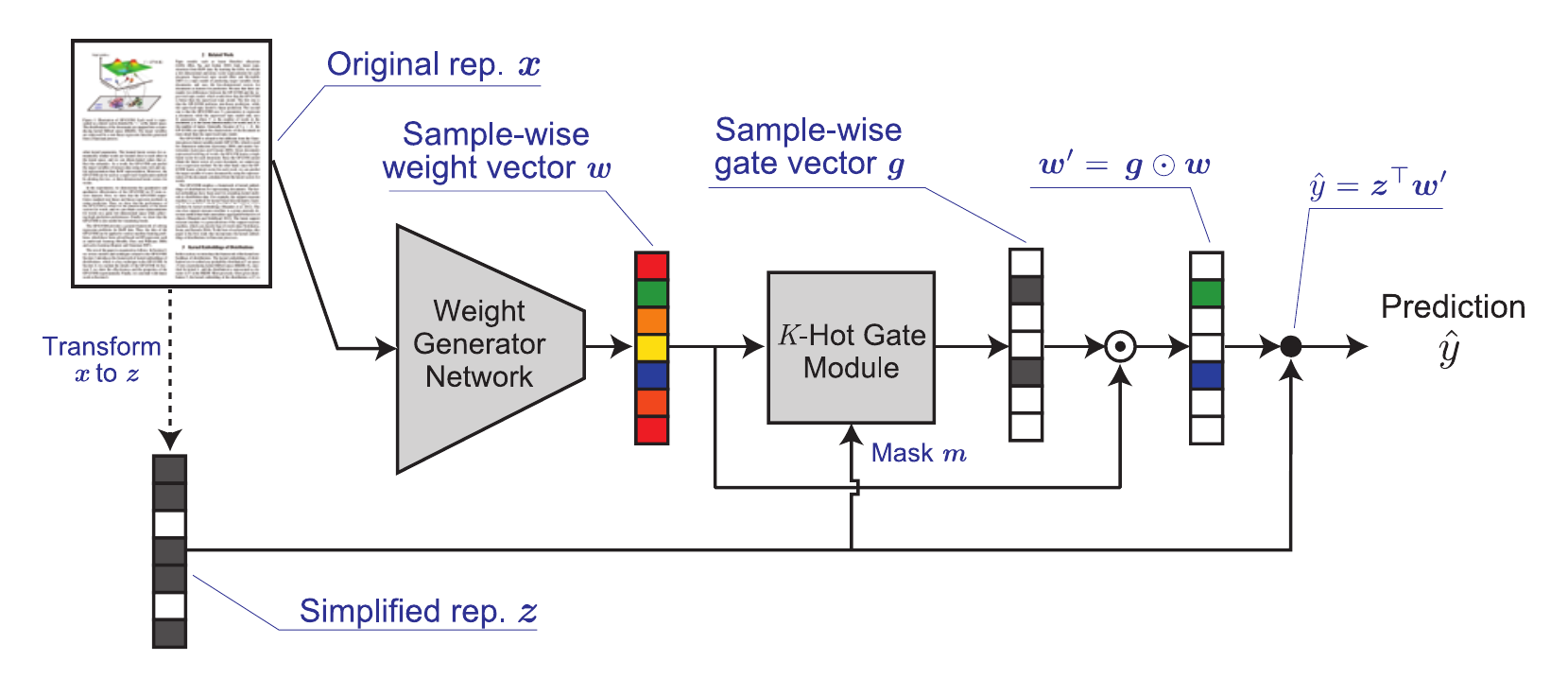}
\caption{
An overview of the NGSLL for binary classification and scalar-value regression.
} 
\label{fig:intro:overview}
\end{figure*}

To combine both the benefits of the high predictive performance of DNNs and high interpretability of linear models into a single model, in this paper we propose
\underline{N}eural \underline{G}enerators of \underline{S}parse \underline{L}ocal \underline{L}inear models (NGSLLs).
The NGSLL generates sparse linear weights for each sample using DNNs, and then makes predictions using the weights.
The overview of the proposed model is illustrated in Figure~\ref{fig:intro:overview}.
The model takes two types of representations as inputs: the original representation (e.g., word sequence)
and its simplified one (e.g., bag-of-words).
For these inputs, it first generates sample-wise dense weights from the original representation using the Weight Generator Network (WGN).
Then, the generated dense weights are fed into the $K$-Hot Gate Module ($K$-HGM) to generate a gate vector that represents $K$ important features and eliminates the weights associated with the remaining irrelevant features.
The dense weights become sparse by the element-wise product with the gate vector.
Finally, it outputs the prediction by the inner product between the sparse weights and the simplified representation.
The parameters of the model are end-to-end trainable via backpropagation.
To be able to robustly learn the parameters when $K$ is set to quite a small value, we learn them based on a coarse-to-fine training procedure.

The principal benefit of the NGSLL is to be able to exploit the rich information in the original representation for the prediction using the simplified representation.
Let us consider sentiment classification on movie review texts, and we are given a sequence of words as the original representation and the binary vector of bag-of-words as the simplified one.
The sentiment polarities for the words are changed by their own contexts.
For example, the polarity of the word ``funny'' should be positive for the context ``This is a {\it funny} comedy'',  and negative in the context of: ``The actor is too bad to {\it funny}''.
Although the difference of the polarities cannot be affected to the weights of ordinary linear models, the NGSLL can capture it by generating sample-wise weights using the WGN.
Additionally, the generated weights are interpretable because they are the parameters of linear models and only a small number of weights are selected by the $K$-HGM.

We demonstrate the effectiveness of the NGSLL by conducting experiments on image and text classification tasks.
The experimental results show that 1) its predictive performance is significantly better than interpretable linear models and the recent models based on neural networks for interpretable machine learning, 2) sample-wise sparse weights generated from the NGSLL are appropriate and helpful to interpret why each of samples is classified as such, and 3) the prediction of NGSLL is computationally more efficient than LIME and SHAP.

\section{Related Work}\label{sec:RW}
In this section, we introduce the existing research related to our own work and explain the differences between our work and them.

In a model-agnostic setting, it is assumed that machine learning models with arbitrary architectures such as DNNs and random forests are acceptable as the prediction models, and that the parameters of the prediction models and their gradients are therefore not available for interpretation.
Thus, the goal of this setting is to explain the prediction model by making use of the relationship between its inputs and the outputs of the prediction model for the inputs.
For example, Chen et al. proposed explaining the prediction models by finding important features in inputs based on mutual information between their response variables and the outputs of the prediction model for those inputs~\cite{chen2018}.

Recently, perturbation-based approach has been well-studied in a model-agnostic setting~\cite{ribeiro2016,lundberg2017,ribeiro2018,fong2017,adler2018}.
The approach attempts to explain the sensitivity of the prediction models to changes in their inputs by using pseudo-samples generated by perturbations.
The representative methods in this approach are LIME~\cite{ribeiro2016} and SHAP~\cite{lundberg2017}.
LIME randomly draws pseudo-samples from a distribution centered at a simplified input to be explained, and fits a linear model to predict the outputs of the prediction model for the pseudo-samples.
SHAP is built upon the idea of LIME, and uses Shapley values to quantify the importance of features of a given input.
However, as described in the introduction, these methods have drawbacks in that the pseudo-samples might be different from the original input distributions, and the learning of the linear model is time-consuming.

Our work considers {\it not} model-agnostic approach, i.e., the prediction and interpretation models can be learned simultaneously or the prediction model itself is interpretable.
As the NGSLL finishes learning its own parameters using only the original input data in the training phase, it can avoid the drawbacks of the perturbation-based approach.

Attentive Mixtures of Experts (AME) is related to our work~\cite{schwab2019}.
With AME, each feature in the input is associated with an expert, and prediction is made by the weighted sum of the outputs of the experts.
By learning the weights by optimizing an objective inspired by Granger-causality, AME executes the selection of important features and the prediction simultaneously.
The functional differences between the NGSLL and AME are 1) the NGSLL can make use of the original representation of inputs, and 2) the NGSLL can specify the number of features used in prediction.

In the computer vision community, explanation methods to highlight the attentive region of an image for its prediction generated by CNNs have been actively studied.
For example, Class Average Mapping (CAM)~\cite{zhou2016} computes channel-wise importance by inserting global average pooling before the last fully-connected layer of CNNs and shows which regions are attentive by visualizing weighted sum of channels with the importance.
Grad-CAM~\cite{selvaraju2017} is a popular gradient-based approach that determines the attentive regions based on the gradients for the outputs of CNNs.
Although these approaches are successfully used in various applications with images, they especially focus on the explanation of CNNs for images.
Unlike these approaches, the NGSLL is a general framework for interpretable machine learning, which can be used independently on the types of data.
Moreover, the NGSLL can incorporate domain-specific knowledge, e.g., the CNNs optimized for images and LSTM-based architectures for natural languages, into the WGN.

Finally, we introduce the relationship between the NGSLL and local linear models which assigns weights to each sample.
Some works proposed to learn them with regularization such that similar samples have similar weights~\cite{hallac2015,yamada2017}.
Here, in the setting of their studies, the similarity between the samples are defined in advance.
The NGSLL can be regarded as employing the regularization implicitly without the similarity, as similar sample-wise weights are generated for similar samples using a shared DNN as the WGN.

\section{Proposed Model}\label{sec:proposed}
In this section, we describe the NGSLL in details.
For sake of explanation, we mainly consider the problem of binary classification, but it can be easily extended for regression and multi-class classification.

Suppose that we are given training data $\set{D} = \{ (\mat{x}_i, \mat{z}_i, y_i) \}_{i=1}^n$ containing $n$ samples.
For ease of explanation, we denote $\mat{x}_i, \mat{z}_i, y_i$ as $\mat{x}, \mat{z}, y$, respectively, by omitting the index.
We assume that two types of inputs are given.
The first one, $\mat{x}$, is the original representation of a sample, which contains rich information representing the sample.
In case of text classification, $\mat{x}$ is for example a sequence of words or characters.
The second one, $\mat{z} \in \mathbb{R}^d$, is a simplified representation for $\mat{x}$ and should be made of the essence of $\mat{x}$ and easy to understand the meaning of each dimension of it.
For example, bag-of-words representation is one of the simplified one for text.
Then, $y \in \{+1, -1\}$ denotes a class label.
Here, if $\mat{z}$ is sparse, mask vector $\mat{m} \in \{0, 1\}^d$ can be used for excluding predefined irrelevant features from $\mat{z}$.
For the case where $\mat{z}$ is a bag-of-words vector, we represent $m_j = 1$ if the $j$th dimension of $\mat{z}$ is zero, that is, its corresponding feature has zero frequency, and $m_j = 0$ otherwise.
In test phase, we assume that both $\mat{x}$ and $\mat{z}$ for a test sample are given.
Our aim is to generate sample-wise sparse linear weights such that high predictive performance and the interpretability of the prediction are archived simultaneously.

Before explaining the details of the NGSLL, we summarize the computation flow of the model from inputs to prediction by reference to Figure~\ref{fig:intro:overview}.
First, the model generates sample-wise weight vector $\mat{w} \in \mathbb{R}^d$ from original representation $\mat{x}$ using Weight Generator Network (WGN).
Then, $\mat{w}$ and mask vector $\mat{m}$ are fed into the $K$-Hot Gate Module ($K$-HGM) to generate gate vector $\mat{g} \in \set{G}$ where $\set{G}$ is defined as
\begin{equation}
\set{G} := \left\{ \mat{g}'\ \Bigg|\ \mat{g}' \in \{0, 1\}^d,\ \mat{m}^\top \mat{g}' = 0,\ \sum_{j=1}^d g'_j = K \right\}.
\end{equation}
Here, $g_j = 1$ indicates that the gate associated with the $j$th dimension of $\mat{g}$ is open.
The role of the $K$-HGM is to leave only $K \in \{ 1, 2, \cdots, d \}$ dimensions in $\mat{z}$ necessary for the prediction and to remove the remaining dimensions.
Using $\mat{w}$ and $\mat{g}$, sample-wise sparse weight vector $\mat{w}' \in \mathbb{R}^d$ is obtained by
\begin{equation}
\mat{w}' = \mat{g} \odot \mat{w},
\label{eq:wprime} 
\end{equation}
where $\odot$ denotes element-wise product operator.
Finally, the prediction $\hat{y}$ for the sample is made by the following equation,  
\begin{equation}
\hat{y} = \mat{z}^\top \mat{w}'.
\label{eq:prediction}
\end{equation}
In the consecutive two subsection, we describe the details of the two key parts, the WGN and the $K$-HGM, in the NGSLL.

\subsection{Weight Generator Network (WGN)}
The WGN is used for generating sample-wise weight vector $\mat{w}$ from the original representation for a sample, $\mat{x}$.
More specifically, extracting features from $\mat{x}$ and generating weights $\mat{w}$ from the features are executed consecutively in the WGN.
Since the feature extraction is a common process for DNNs, we can reuse the existing network architecture and its pretrained model that are good for solving the supervised learning task related to what we want to solve.
Then, in order to transform the features to $\mat{w}$, we stack $L$ fully-connected (FC) layers on the network for the feature extraction.

\subsection{$K$-Hot Gate Module ($K$-HGM)}
\begin{figure}[t]
\centering
\includegraphics[width=120mm,pagebox=artbox]{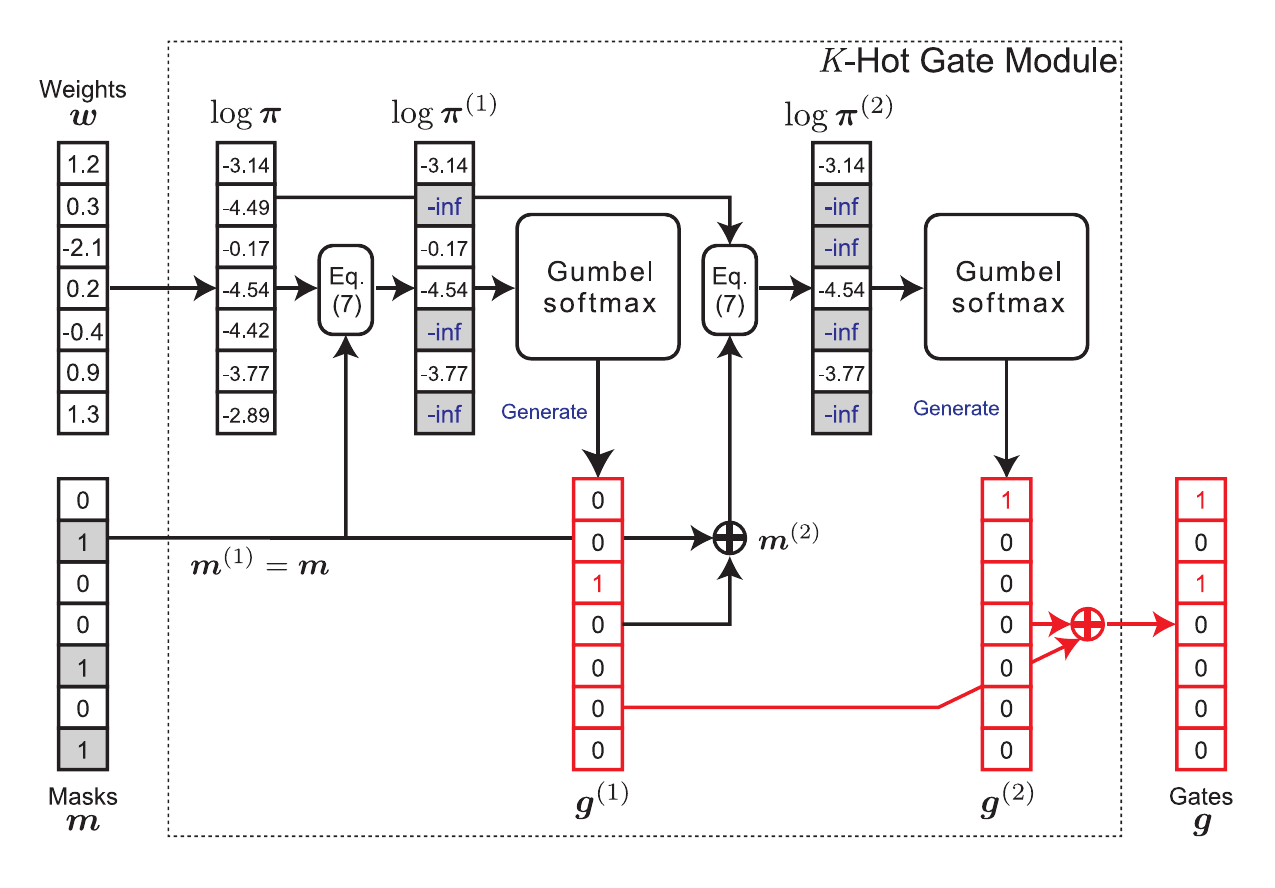}
\caption{A behavior of $K$-Hot Gate Module ($K$-HGM) with $K=2$.
For two inputs, weight vector $\mat{w}$ and mask vector $\mat{m}$, the logarithm of probability vector $\log\mat{\pi}$ is calculated first.
Then, $\log\mat{\pi}^{(1)}$ is obtained using $\log\mat{\pi}$ and $\mat{m}^{(1)}$ based on~\bref{eq:logpi}, and the first gate vector $\mat{g}^{(1)}$ is sampled from Gumbel softmax with $\log\mat{\pi}^{(1)}$ based on~\bref{eq:gumbel_softmax}.
Using $\mat{m}^{(1)}$ and $\mat{g}^{(1)}$,  $\mat{m}^{(2)}$ is updated based on~\bref{eq:m}.
This process is repeated until $K$ gate vectors are generated, and finally $K$-HGM outputs $K$-hot gate vector $\mat{g}$ by summing up them.
}
\label{fig:proposed:KHGM}
\end{figure}

A naive solution for making prediction using only $K$ important features is to obtain the top-$K$ values in the absolute values of the generated weight vector, $|\mat{w}|$, and then use only their weights in \bref{eq:prediction}.
However, this solution does not necessarily select important features correctly, because it prevents learning models in an end-to-end manner. 
To avoid this drawback, the NGSLL uses the $K$-HGM.
Figure~\ref{fig:proposed:KHGM} illustrates an example for the behavior of the $K$-HGM with $K=2$.

The $K$-HGM is given two types of inputs, weight vector $\mat{w}$ generated by the WGN and mask vector $\mat{m}$.
When finding important features using $\mat{w}$, its absolute values are meaningful.
However, since the absolute function is not differentiable at zero, we use its element-wise square value $\mat{w}^2$ instead.
To generate $K$-hot gate vector $\mat{g} \in \mathcal{G}$ such that the dimensions associated with the $K$ largest values in $\mat{w}^2$ are one, indifferentiable $\argmax$ operation is needed.
However, this prevents end-to-end training of the proposed model via backpropagation.
To avoid this drawback, we employ Gumbel softmax~\cite{jang2016}, a continuous approximation for $\argmax$.
We generate $\mat{g}$ by sequentially sampling $K$ one-hot vectors $\mat{g}^{(1)}, \mat{g}^{(2)}, \cdots, \mat{g}^{(K)}$ using Gumbel softmax and then summing up them, that is,
\begin{equation}
\mat{g} = \sum_{t=1}^K \mat{g}^{(t)}.
\end{equation}
Here, the value of the $j$th element for the $t$th gate vector, $g^{(t)}_j$, can be calculated as follows:
\begin{equation}
g^{(t)}_j = \frac{\exp\left(\left(\log\pi_j^{(t)} + \lambda_j\right) / \tau \right)}{\sum_{j'=1}^d \exp\left(\left(\log\pi_{j'}^{(t)} + \lambda_{j'}\right) / \tau \right)},
\label{eq:gumbel_softmax}
\end{equation}
where $\tau > 0$ is a temperature parameter and $\lambda_j \in \mathbb{R}$ is a sample drawn from the standard Gumbel distribution, which can be obtained by the following processes:
\begin{equation}
u_j \sim \mathrm{Uniform(0, 1)}, \quad \lambda_j = -\log\left(-\log(u_j)\right).
\end{equation}

As shown in~\bref{eq:gumbel_softmax}, Gumbel softmax requires the logarithm of a normalized $d$-dimensional probability vector.
In the NGSLL, we calculate it as $\log\mathrm{softmax}(\mat{w}^2)$.
Then, the model changes it over $t \in \{1, 2, \cdots, K\}$ to prevent the same elements in the already generated one-hot gate vectors from being one multiple times.
We denote the $t$th one as $\log\mat{\pi}^{(t)}$ whose the $j$th element is calculated as follows:
\begin{equation}
\log\pi^{(t)}_j = \begin{cases}
\log\pi_j & ( m^{(t)}_j = 0 ) \\
-\infty & ( m^{(t)}_j = 1 ),
\end{cases}
\label{eq:logpi}
\end{equation}
where $\log\pi^{(t)}_j = -\infty$ leads to be $g_j^{(t)} = 0$ in~\bref{eq:gumbel_softmax} because of $\exp(-\infty) = 0$.
Here, $m^{(t)}_j \in \{0, 1\}$ indicates whether the $j$th elements in $\mat{g}^{(1)}, \mat{g}^{(2)}, \cdots, \mat{g}^{(t-1)}$ are already being one ($m^{(t)}_j = 1$) or not ($m^{(t)}_j = 0$).
More specifically, given the $t$th mask vector $\mat{m}^{(t)}$ and gate vector $\mat{g}^{(t)}$, the next mask vector $\mat{m}^{(t+1)}$ is obtained by
\begin{equation}
\mat{m}^{(t+1)} = \mat{m}^{(t)} + \mathtt{onehot}\left(\argmax \mat{g}^{(t)} \right),
\label{eq:m}
\end{equation}
where we define $\mat{m}^{(1)} = \mat{m}$, and $\mathtt{onehot}(j) \in \{0, 1\}^d$ is a one-hot vector being one only at the $j$th element and zero at the other elements.

\subsection{Coarse-to-Fine Training}
With the NGSLL, the parameters to be learned are only those of WGN and their estimation can be done by minimizing the cross entropy loss between true label $y$ and predicted one $\hat{y}$ in a standard supervised way for deep learning.

However, when $K$ is set to a quite small value, e.g., $K=1$, an ordinary training often did not work well in our preliminary experiments.
This is because the gradients for the dimensions having zero in gate vector $\mat{g}$ are not propagate to the WGN.
More specifically, in the computation of the backpropagation, we need the gradient vector for \bref{eq:wprime} in terms of parameter $\theta \in \mathbb{R}$ of the WGN, which is calculated as follows:
\begin{equation}
\frac{\partial \mat{w}'}{\partial \theta}
= \frac{\partial \mat{w}'}{\partial \mat{w}} \frac{\partial \mat{w}}{\partial \theta}
= \mat{g} \odot \frac{\partial \mat{w}}{\partial \theta}.
\end{equation}
From this equation, the values of the gradient vector are obviously determined by $\mat{g}$, and the dimensions with zero in $\mat{g}$ also become zero in the gradient vector.
Due to this, with getting $K$ smaller, many parameters cannot be updated enough and the learning will be converged at a poor local optimum.
We overcome this difficulty by employing a coarse-to-fine training procedure in which coarse training phase and fine one are executed in a step-by-step manner.

In the coarse training phase, we first set $K$ to a high value, e.g., $K=10$, and temperature parameter $\tau$ for Gumbel softmax~\bref{eq:gumbel_softmax} to a relatively large value, e.g., $\tau = 1$.
Here, with getting $\tau$ larger, the gate vector $\mat{g}$ is closer to a sample from an uniform distribution.
By learning the parameters under such a setting, we can prevent the emergence of zero gradients.
As a result, the parameters can be sufficiently updated, although the effect of the $K$-HGM is barely affected to the parameters yet.

After the loss is converged in the coarse training phase, we move to the fine training phase.
We reset $K$ and $\tau$ to one's desired values, and then restart to learn the model from the converged point in the coarse training phase.

\section{Experiments}\label{sec:experiment}
In this section, we show the experimental results of image and text classification tasks.
The aim of the experiments is to show that: 1) the NGSLL can achieve high predictive performance; 2) the sparse weights generated by the NGSLL are appropriate and helpful for the interpretation of prediction; and 3) the NGSLL is computationally efficient in the test phase.
All the experiments were done with a computer with Intel Core i9 7900X 3.30GHz CPU, NVIDIA GeForce RTX2080Ti GPU, and 64GB of main memory.

\paragraph{Setup of the NGSLL.}
The hyperparameters of the NGSLL to be tuned are the number of FC layers connecting between the WGN and the $K$-HGM, $L$, the number of units for the FC layers, $U$, and temperature parameter for Gumbel softmax, $\tau$.
With $L$ and $U$, we assign their values from $L \in \{1, 2\}$ and $U \in \{128, 256\}$, we choose the optimal ones which achieved the best accuracy on the validation set for each dataset.
In the experiments we set $\tau$ to $1$ and $0.1$ in the coarse and fine training, respectively.
The parameters of the NGSLL are optimized by Adam~\cite{kingma2014} with the default hyperparameters in the coarse training and Momentum SGD~\cite{qian1999} with a decay rate of the first order moment, $p=0.9$, in the fine training.
Here, we set the learning rate of the Momentum SGD to one tenth of the final learning rate of Adam.

\begin{table}[t]
\centering
\caption{Predictive accuracy on binary-class MNIST dataset. 
Numbers in brackets are their standard deviations.
Note that, the accuracy of AME and DNN is constant to $K$.
}
\label{tab:experiment:mnist_acc}
\begin{tabular}{@{}rrrr@{}}
\toprule
                                                              & $K=1$                & $K=5$                & $K=10$               \\ \midrule
NGSLL                                                         & {\bf 0.994} ($\pm{0.000}$) & {\bf 0.995} ($\pm{0.001}$) & {\bf 0.994} ($\pm{0.000}$) \\
\begin{tabular}[c]{@{}r@{}}NGSLL\\ (w/o $K$-HGM)\end{tabular} & 0.967 ($\pm{0.009}$) & 0.991 ($\pm{0.001}$) & 0.991 ($\pm{0.001}$) \\
Ridge                                                         & 0.526 ($\pm{0.000}$) & 0.561 ($\pm{0.000}$) & 0.582 ($\pm{0.002}$) \\
Lasso                                                         & 0.523 ($\pm{0.005}$) & 0.549 ($\pm{0.001}$) & 0.582 ($\pm{0.003}$) \\ \midrule
AME ($\alpha=0$)                                              & \multicolumn{3}{c}{0.973 ($\pm{0.003}$)}                           \\
AME ($\alpha=.05$)                                            & \multicolumn{3}{c}{0.977 ($\pm{0.001}$)}                           \\ 
DNN                                                           & \multicolumn{3}{c}{0.992 ($\pm{0.001}$)}                           \\ \bottomrule
\end{tabular}
\end{table}

\paragraph{Comparing models.}
We compare the NGSLL with various interpretable models such as Lasso~\cite{tibshirani1996}, Ridge~\cite{hoerl1970}, AME~\cite{schwab2019}, LIME~\cite{ribeiro2016} and SHAP~\cite{lundberg2017}.
Lasso and Ridge are linear models with $\ell_1$ and $\ell_2$ regularization, respectively.
Unlike the NGSLL, the weights of their models are common across all the samples.
We use scikit-learn\footnote{\url{https://scikit-learn.org/stable/}} for their implementations, and their hyperparameter, the strength of regularization, is optimized by cross validation.
AME is a method for estimating feature importance with neural networks based on the attentive mixture of experts.
We use the authors' implementation\footnote{\url{https://github.com/d909b/ame}} for the MNIST dataset with a small change.
AME can control the strength of the regularization for the feature importance by $\alpha \in [0, 1]$.
As with the original paper on AME, we set $\alpha$ to 0 and 0.05.
LIME and SHAP are model-agnostic methods and learn sample-wise weights for each test sample using the predictions for its perturbed samples.
We use the authors' implementations for LIME\footnote{\url{https://github.com/marcotcr/lime}} and SHAP\footnote{\url{https://github.com/slundberg/shap}}.
As these methods do not produce prediction models, we use them for comparing computational efficiency in the test phase.
For reference, we also compare the NGSLL with standard approaches using DNN with original representation as input.
The DNN architecture is almost the same as the WGN of the NGSLL, but its last FC layer directly outputs label probabilities instead of sample-wise weights.
Since the DNN is not interpretable, it is inappropriate for our task, but it can be regarded as the upper bound for the predictive performance of the NGSLL.
To evaluate the impact of the $K$-HGM, we also compare it with the NGSLL without $K$-HGM.
As it generates sample-wise weights through the $K$-HGM, the weights become dense.
With Lasso, Ridge and the NGSLL without $K$-HGM, evaluating accuracy at $K$ is performed by using weights with the largest $K$ absolute values for the weights.

\subsection{Handwritten Digits Image Classification}
\paragraph{Dataset preparation.}
In this experiments, we use MNIST~\cite{lecun1998}, a standard benchmark dataset for handwritten digits classification.
Although the original MNIST is made of 10 classes corresponding to digits from 0 to 9, we split the classes into two groups, i.e., we reassign $y=-1$ for the images with digits 0 to 4 and $y=1$ for the images with digits 5 to 9.
This is to show that the NGSLL can generate sparse weights appropriate for each image, because the weights should be different even among the images in the same class due to large changes in their digits.
We call this {\it binary-class MNIST}.
The original MNIST contains 60,000 training images and 10,000 test images.
With the binary-class MNIST, we use 5,000 images randomly selected from the training images as validation images, and the remaining 55,000 images as training images.
In such a fashion, we create five training and validation sets with different random seeds to evaluate the robustness of the NGSLL.
We use the original test images for evaluation.
The images are gray-scale and their resolution is 28 $\times$ 28.
Although each pixel of the images has an integer value of 0 to 255, we transform it to $[0, 1]$ and use the image as original representation $\mat{x}$.
Its simplified representation is constructed by downsampling $\mat{x}$ to $7 \times 7$ and converting it to vector $\mat{z} \in \mathbb{R}^{49}$.

\paragraph{WGN architecture.}
We use a simple three-layer CNN architecture as the WGN.
Each layer is constructed by a sequence of convolution, ReLU and max pooling modules.
The same architecture is used in DNN for fair comparison.
After the output of the third layer in the WGN, it is converted to weight vector $\mat{w} \in \mathbb{R}^{49}$ via FC layers.

\paragraph{Predictive performance.}

\begin{figure*}[t]
\centering
\includegraphics[width=\textwidth,pagebox=artbox]{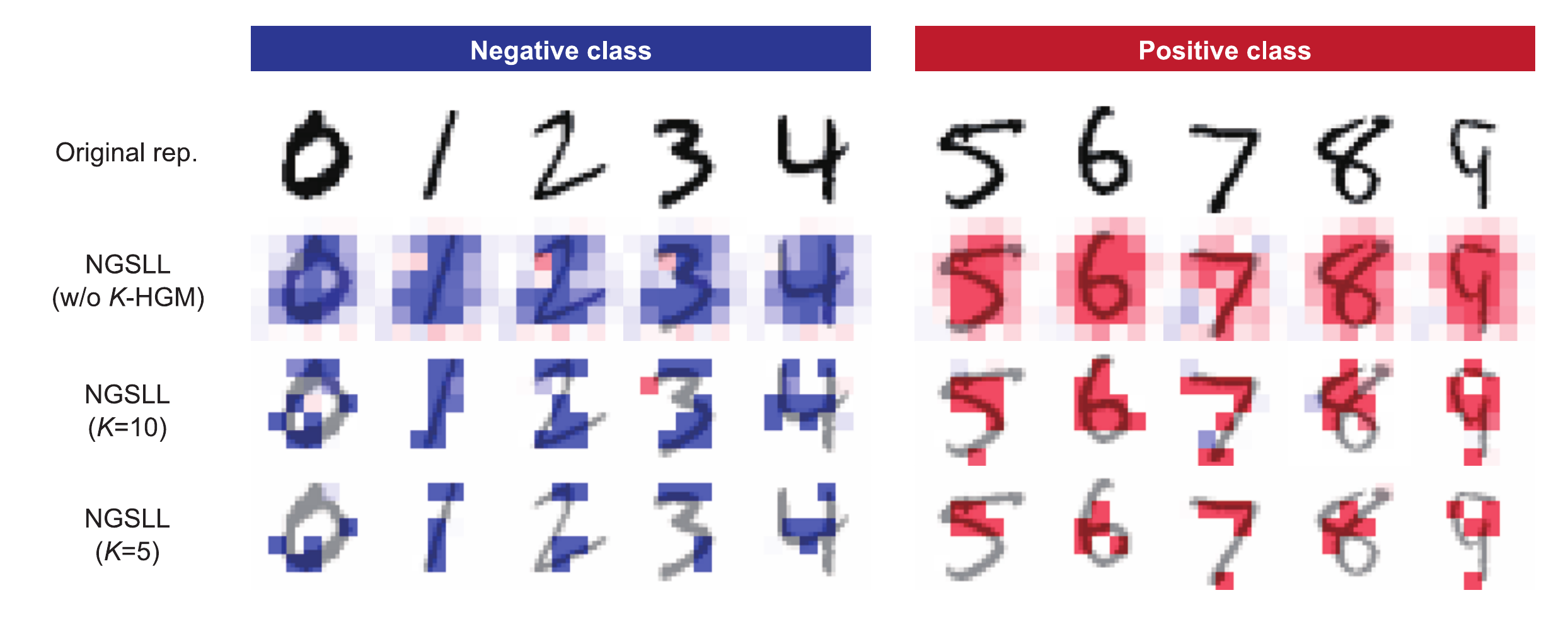}
\caption{
Weights visualization for binary-class MNIST dataset.
Red and blue colors denote positive and negative weights, respectively, and their color strength represents the magnitude of the weights.
} 
\label{fig:experiment:mnist_visualization}
\end{figure*}

\begin{table}[t]
\centering
\caption{
The average running times [ms] taken to interpretation for a sample on binary-class MNIST dataset.
Numbers in brackets are their standard deviations.
Note that the running time taken to prediction for a sample by the prediction model (random forests) used in SHAP and LIME is 0.6 ms in average.
The times of SHAP and LIME are constant to $K$.
}
\label{tab:experiment:mnist_time}
\begin{tabular}{@{}rcrr@{}}
\toprule
      & \multicolumn{1}{r}{$K=1$} & $K=5$ & $K=10$ \\ \midrule
NGSLL & \multicolumn{1}{r}{{\bf 2.55} ($\pm{0.04}$)}  & {\bf 6.19} ($\pm{0.07}$)  & {\bf 10.76} ($\pm{0.17}$)  \\
SHAP  & \multicolumn{3}{c}{699.65 ($\pm{60.39}$)}                 \\
LIME  & \multicolumn{3}{c}{1109.95 ($\pm{22.11}$)}                \\ \bottomrule
\end{tabular}
\end{table}

Table~\ref{tab:experiment:mnist_acc} shows the predictive performance of each model on the binary-class MNIST dataset.
The table presents the following three facts.
First of all, the NGSLL outperforms the other models regardless of the $K$ values.
As the accuracy of Ridge and Lasso shows, it is very hard to solve this task by such ordinary linear models with the simplified representation.
Although the NGSLL also uses the simplified representation in prediction~\bref{eq:prediction}, it can achieve high accuracy by using sample-wise weights generated by the WGN.
Second, the accuracy of the NGSLL is comparable to that of the DNN.
This implies that the NGSLL can effectively transfer useful information included in the original representation into its simplified one, without information loss.
Third, even when $K=1$, the NGSLL can achieve high accuracy, although the accuracy of the NGSLL without $K$-HGM decreases.
This is because the NGSLL can learn its parameters optimized to the $K$ value due to employing the $K$-HGM.

\paragraph{Weights visualization.}
Figure~\ref{fig:experiment:mnist_visualization} shows the weights generated by the NGSLL for the image of each digit.
For each image in binary-class MNIST, it is important for correct prediction to assign weights with the sign of its label to black pixels in its original representation.
With the NGSLL without $K$-HGM, the weights with the correct sign can be generated.
However, it is difficult to interpret which regions in the image are important for prediction since large weights are assigned to many regions.
On the other hand, the NGSLL in cases of both $K=5$ and $10$ can appropriately generate sparse weights only on black pixels by capturing the shape of the digits displayed in the images.
Thanks to the sparse weights, one can easily understand which regions in the image are useful in prediction.

\paragraph{Computational time.}
Table~\ref{tab:experiment:mnist_time} shows the average running times taken to interpretation for a sample.
Note again that, as SHAP and LIME are model-agnostic, they learn a linear model for interpretation by making use of the outputs of the given prediction model for the perturbed samples of the given sample.
Here, the prediction model for SHAP and LIME in this experiment is random forests, and its running time for a sample is 0.6 ms in average.
As shown in the table, the NGSLL can generate linear weights for interpretation about 270-430 times faster than SHAP and LIME.
We also found that the running times of the NGSLL increase with increasing the $K$ values.
This is because the $K$-HGM generates $K$ one-hot vectors repeatedly, and therefore, the times increase linearly with respect to $K$.

\subsection{Text Classification}

\begin{figure*}[t]
\begin{minipage}{0.3\hsize}
\begin{center}
\includegraphics[width=49mm,pagebox=artbox]{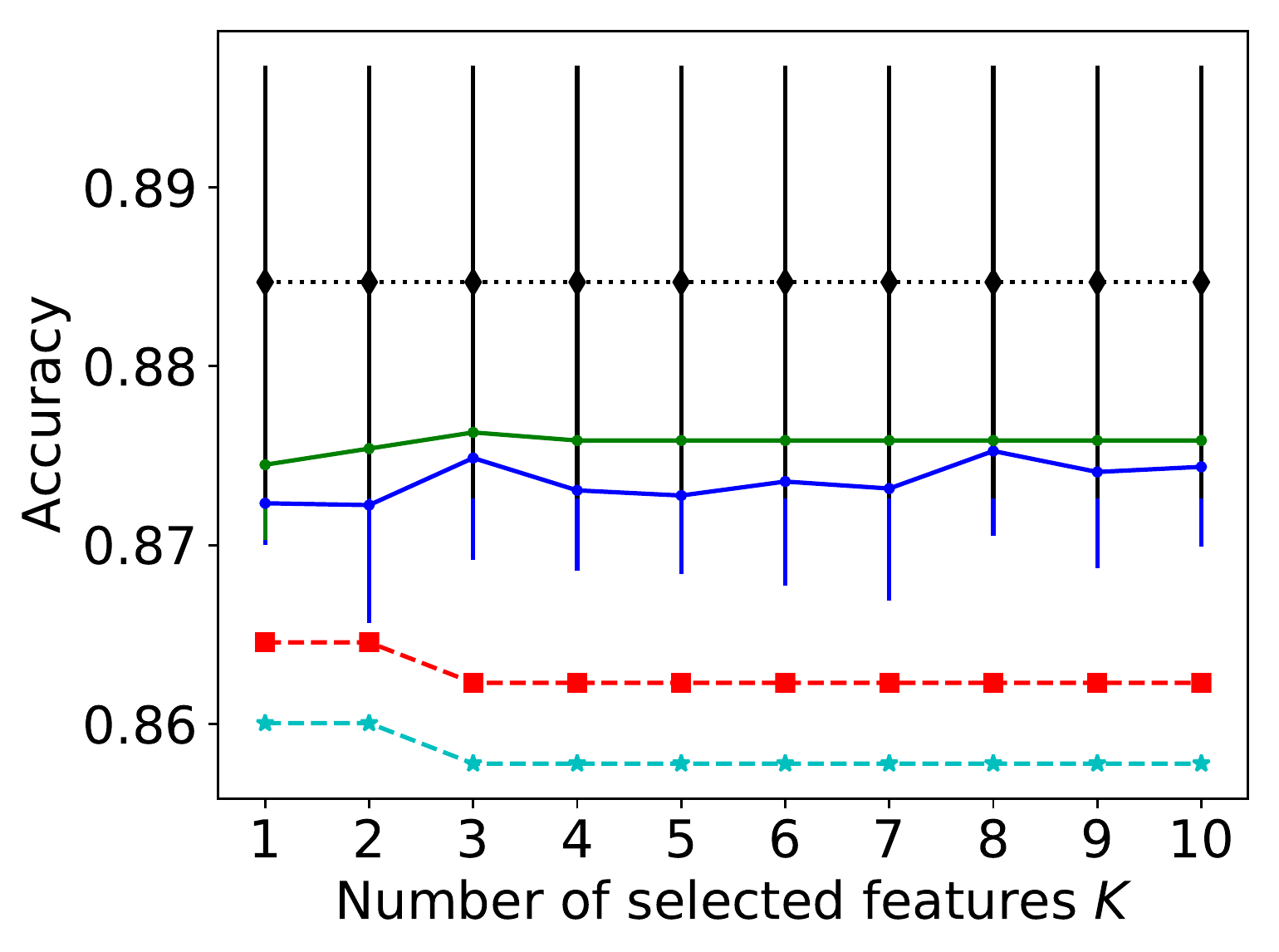}\\(a) MPQA
\end{center}
\end{minipage}
\begin{minipage}{0.3\hsize}
\begin{center}
\includegraphics[width=49mm,pagebox=artbox]{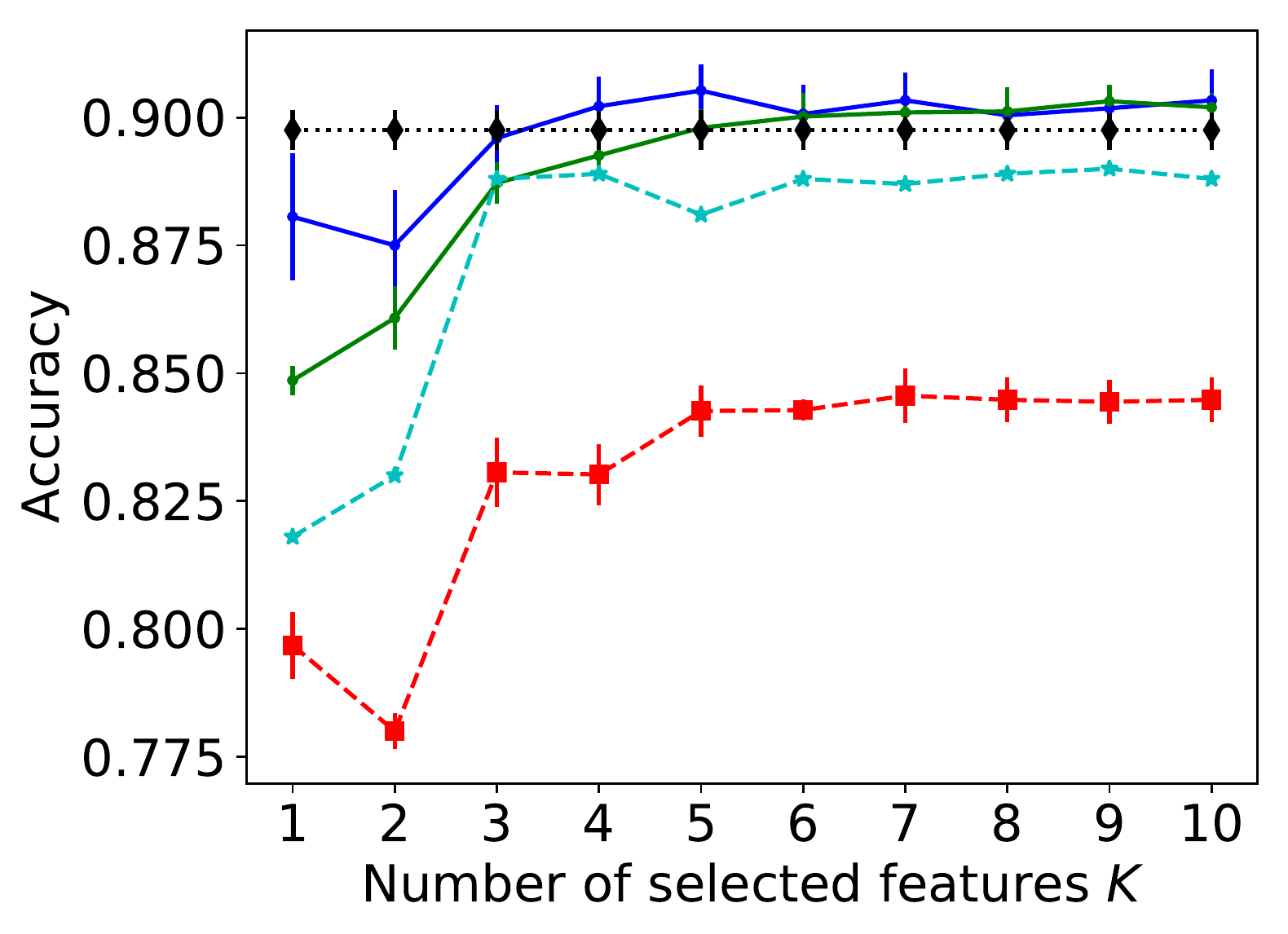}\\(b) Subj
\end{center}
\end{minipage}
\begin{minipage}{0.39\hsize}
\begin{center}
\includegraphics[width=70mm,pagebox=artbox]{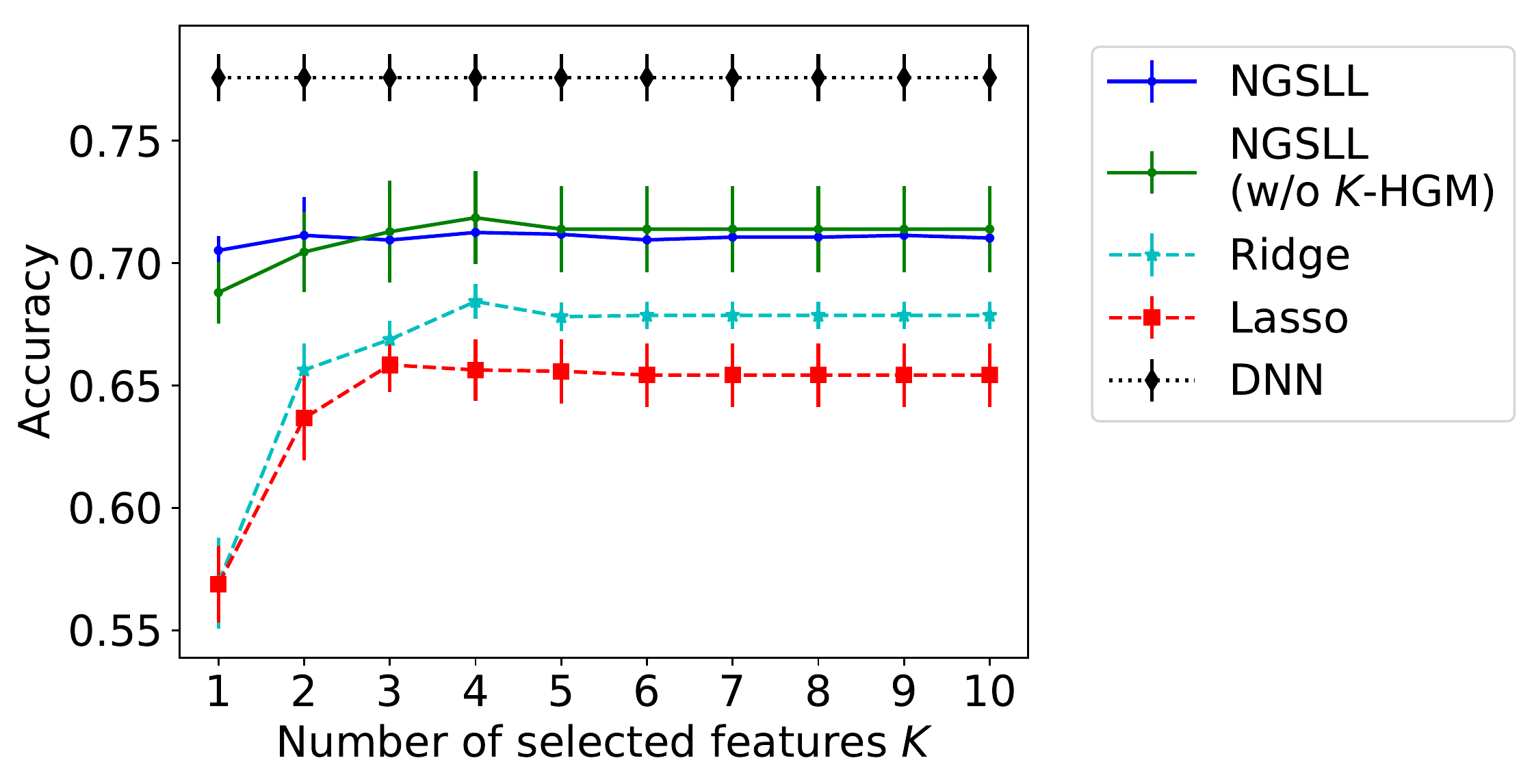}\\(c) TREC
\end{center}
\end{minipage}
\caption{Predictive accuracy on text classification datasets.
The error bar at each point denotes its standard deviation.
Note that the accuracy of DNN can be regarded as the upper bound for the NGSLL and is constant to $K$.
}
\label{fig:experiment:text_accuracy}
\end{figure*}

\begin{figure*}[t]
\centering
\vspace{5mm}
\includegraphics[width=\textwidth,pagebox=artbox]{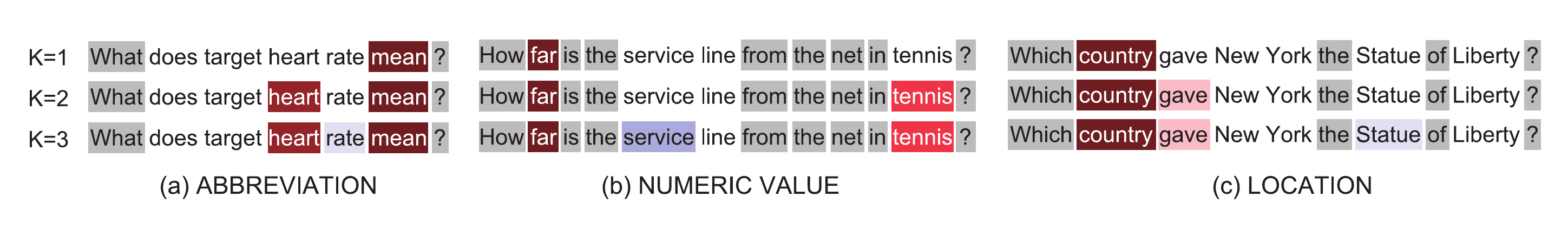}
\caption{
Weights visualization for the sentences with three answer types (a)-(c) on the TREC dataset.
Red and blue colors denote positive and negative weights, respectively, and their color strength represents the magnitude of the weights.
Grey color denotes out-of-vocabulary words.
} 
\label{fig:experiment:text_visualization}
\end{figure*}

\paragraph{Dataset preparation.}
In this experiments, we use three text classification datasets, referred to as MPQA~\cite{wiebe2005}, Subj~\cite{pang2002} and TREC~\cite{li2002}.
The MPQA is a dataset to predict its opinion polarity (positive or negative) from a phrase.
It contains 10,603 phrases in total.
The Subj is a dataset to predict subjectivity (subjective or objective) from a sentence about a movie.
It contains 10,000 sentences in total.
The TREC is a dataset to predict the type of its answer from a factoid question sentence.
It contains 5,952 sentences in total, and each sentence has one out of six answer types: abbreviation, entity, description, location and numeric value.
With each of the datasets, we create five sets by randomly splitting it into training, validation and test parts.
We in advance eliminate stopwords and the words appearing only once in each of the datasets.
As a result, the vocabulary size of the MPQA, the Subj and the TREC is 2,661, 9,956 and 3,164, respectively.
Each sentence and phrase is used as a word sequence of it for the original representation, while a bag-of-words vector of it for the simplified one.
Here, in the case where unknown words appear in the word sequence, we replace it with a special symbol.

\paragraph{WGN architecture.}
We use sentence CNN~\cite{kim2014}, a popular and standard CNN architecture used for sentence classification, as the WGN.
All the parameters of the WGN are learned from a scratch.
As with the experiments for image classification, the same architecture is used in DNN.

\paragraph{Predictive performance.}
Figure~\ref{fig:experiment:text_accuracy} shows the accuracy of each model with varying $K$ on the three text datasets.
We obtained similar findings with those on the binary-class MNIST in the following two points of view.
First, the NGSLL outperforms Ridge and Lasso regardless of the $K$ values.
This indicates that the WGN in the NGSLL takes advantage of the information of word sequences for accurate prediction.
Secondly, the NGSLL can maintain high accuracy even when $K$ becomes a small value, as shown especially in the Subj and the TREC.
Conversely, the different result with the those on the binary-class MNIST is that the accuracy of the NGSLL is lower than the one of DNN on the MPQA and the TREC, while their accuracy are almost the same on the Subj.
Although the reason is not trivial, this result suggests that the accuracy gap between the NGSLL and DNN would emerge depending on dataset choice and WGN architecture.

\paragraph{Weights visualization.}
Figure~\ref{fig:experiment:text_visualization} shows the visualization results of the sparse weights generated by the NGSLL with K-HGM for the sentences on the TREC.
As the TREC is for the task of predicting an answer type from a sentence, large weights should be assigned to words suitable for predicting the answer type.
As shown in the results when $K=1$, the NGSLL can find the representative word for each answer type, e.g., ``far'' for answer type (b) numerical value.
Although irrelevant words for the answer type tend to be remained even when the $K$ value becomes large, the representative word is consistently selected as an important one.

\section{Conclusion}\label{sec:conclusion}
In this paper we have proposed neural generators of sparse local linear models (NGSLLs), where we bring both the benefits of high predictive performance of DNNs and high interpretability of linear models.
The NGSLL generates sparse and interpretable linear weights for each sample using a DNN, and then makes prediction using the weights.
Here, the weights associated with only $K$ important features become large and the remaining weights shrink to zero by using our gating module based on Gumbel softmax.
According to the experimental results on image and text classification tasks, we have found that 1) the NGSLL can robustly achieve high accuracy regardless of the $K$ values, 2) the NGSLL can provide reasonable evidences for prediction by visualizing the weights, and 3) the prediction by the NGSLL is computationally efficient.

\section*{Acknowledgment}
A part of this work was supported by JSPS KAKENHI Grant Number JP18K18112.

\bibliography{zotero} 
\bibliographystyle{unsrt}

\end{document}